\documentclass[10pt,twocolumn,letterpaper]{article}
\usepackage{cvpr} 
\usepackage[pagebackref,breaklinks,colorlinks]{hyperref}
\usepackage[capitalize]{cleveref}
\usepackage{times}
\usepackage{epsfig}
\usepackage{graphicx}
\usepackage{amsmath,amssymb}
\usepackage{booktabs}
\usepackage{multirow}
\usepackage{array}
\usepackage{caption}

\def\confName{CVPR}
\def\confYear{2026}

\begin{document}
\title{Multimodal Emotion Regression with Multi-Objective Optimization and VAD-Aware Audio Modeling for the 10th ABAW EMI Track}

\author{
\begin{tabular}{ccc}
Jiawen Huang$^{1,2}$ & Chenxi Huang$^{1,2}$ & Zhuofan Wen$^{1,2}$ \\
Hailiang Yao$^{3,2}$ & Shun Chen$^{1,2}$ & Longjiang Yang$^{1,2}$ \\
Cong Yu$^{1,2}$ & Fengyu Zhang$^{1,2}$ & Ran Liu$^{3,2}$ \\
Bin Liu$^{2}$  & \\[0.5em]
\end{tabular}\\[0.5em]
$^{1}$School of Artificial Intelligence, University of Chinese Academy of Sciences \\
$^{2}$Institute of Automation, Chinese Academy of Sciences \\
$^{3}$Tianjin Normal University \\
[0.5em]
{\tt\small
huangjiawen25@mails.ucas.ac.cn, huangchenqian22@mails.ucas.ac.cn, wenzhuofan2023@ia.ac.cn,} \\
{\tt\small
2311090059@stu.tjnu.edu.cn, chenshun2023@ia.ac.cn, yanglongjiang2024@ia.ac.cn,} \\
{\tt\small
yucong2024@ia.ac.cn, zhangfengyu2024@ia.ac.cn, 2175278397@qq.com,} \\
{\tt\small
liubin@nlpr.ia.ac.cn}
}
\maketitle
\begin{abstract}
We participated in the 10th ABAW Challenge, focusing on the Emotional Mimicry Intensity (EMI) Estimation track on the Hume-Vidmimic2 dataset. This task aims to predict six continuous emotion dimensions: Admiration, Amusement, Determination, Empathic Pain, Excitement, and Joy. Through systematic multimodal exploration of pretrained high-level features, we found that, under our pretrained feature setting, direct feature concatenation outperformed the more complex fusion strategies we tested. This empirical finding motivated us to design a systematic approach built upon three core principles: (\textit{i}) preserving modality-specific attributes through feature-level concatenation; (\textit{ii}) improving training stability and metric alignment via multi-objective optimization; and (\textit{iii}) enriching acoustic representations with a VAD-inspired latent prior. Our final framework integrates concatenation-based multimodal fusion, a shared six-dimensional regression head, multi-objective optimization with MSE, Pearson-correlation, and auxiliary branch supervision, EMA for parameter stabilization, and a VAD-inspired latent prior for the acoustic branch. On the official validation set, the proposed scheme achieved our best mean Pearson Correlation Coefficient of \textbf{0.478567}.
\end{abstract}

\section{Introduction}
Emotional mimicry refers to the spontaneous tendency of an individual to reflect or align with the affective expressions of an interaction partner. As an important component of social cognition, empathy, and interpersonal communication, emotional mimicry has attracted growing attention in affective computing and human-centered artificial intelligence. Automatically estimating emotional mimicry intensity from multimodal behavioral signals can benefit a range of applications, including socially aware dialogue systems, mental health assessment, and computational analysis of human interaction.

The Emotional Mimicry Intensity (EMI) Estimation track of the 10th ABAW Challenge formulates this problem as six-dimensional continuous emotion regression from synchronized visual, acoustic, and textual information. Compared with categorical emotion recognition, EMI estimation is considerably more challenging. The target is continuous, weakly expressed, and often entangled with speaker-specific style, contextual semantics, and cross-modal asynchrony. Therefore, the key challenge lies not only in extracting informative modality-specific cues, but also in integrating them in a way that is both robust and compatible with the evaluation objective.
\begin{figure*}[htbp]
    \centering
    \includegraphics[width=\textwidth]{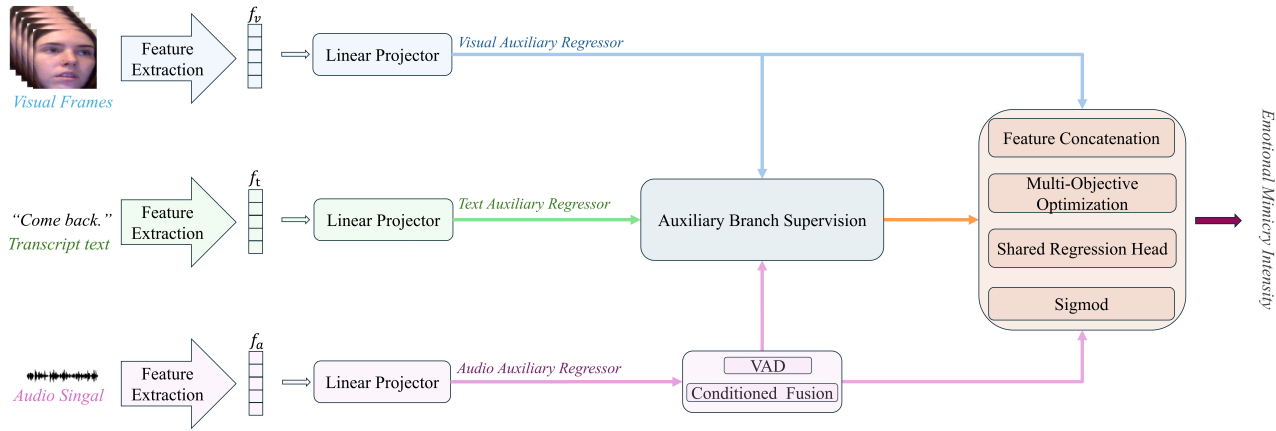}
    \caption{Architecture of the proposed multimodal framework. The model employs three modality-specific branches with auxiliary supervision, VAD-Aware audio fusion, feature concatenation, and a shared six-dimensional regression head for emotional mimicry intensity estimation.}
    \label{fig:1}
\end{figure*}
A prevailing trend in multimodal affective modeling is to strengthen temporal and cross-modal interaction modeling through recurrent networks, temporal convolutions, or multimodal Transformers. While such designs can improve representational power, their benefit is not always guaranteed when strong pretrained features are already available. In our setting, we observe that explicitly introducing additional temporal modeling on top of pretrained multimodal features does not improve the final result and can even introduce side effects in optimization. This finding suggests that, under our pretrained feature pipeline, the performance bottleneck may lie less in additional temporal remodeling and more in fusion and optimization design.

Motivated by this observation, we revisit the design space of multimodal emotion regression and focus on three components. First, we use feature concatenation as the final fusion strategy, which preserves the heterogeneity of visual, audio, and text representations and consistently outperforms average fusion in our experiments. Second, we adopt a multi-objective optimization scheme that combines mean squared error with batch-level Pearson correlation loss, together with auxiliary branch-level supervision, to better align training with the challenge metric and improve optimization behavior. Third, we propose a VAD-Aware audio modeling mechanism that injects affect-inspired information into the acoustic branch through a latent three-dimensional prior, thereby strengthening the contribution of audio to multimodal regression.

Based on these designs, our final system achieves an average Pearson correlation of 0.4786 on the official validation split. Our study provides an empirical perspective on EMI estimation: when high-quality pretrained multimodal features are available, careful fusion and objective design may be more critical than introducing additional temporal modules.

The contributions of this work are threefold:
\begin{itemize}[leftmargin=1.5em]
    \item \textbf{A feature-concatenation multimodal regression framework.} We demonstrate that, given strongly pretrained unimodal features, a simple concatenation-based fusion strategy outperforms more complex alternatives. Our framework preserves heterogeneous modality-specific information and enables unified prediction through a shared regression head, establishing a strong and computationally efficient baseline for EMI estimation.
    \item \textbf{A multi-objective optimization and stable training protocol.} We introduce a composite objective that combines MSE supervision for pointwise accuracy, a Pearson-correlation loss for metric alignment, and auxiliary branch supervision for branch-wise representation learning, together with EMA for parameter smoothing. This multi-faceted optimization strategy enhances both training stability and generalization performance.
    \item \textbf{A VAD-Aware audio modeling module.} We enrich acoustic representations with a VAD-inspired latent pathway that serves as an affect-aware regularizer. This auxiliary module provides the acoustic branch with a low-dimensional affect-inspired prior, yielding consistent and substantial gains in multimodal emotion regression across all six target dimensions.
\end{itemize}

We evaluate our proposed framework on the official validation split of the Hume-Vidmimic2 dataset. Experiments demonstrate that our approach consistently outperforms both simple fusion baselines and more elaborate alternatives, achieving an average Pearson correlation of 0.4786. Ablation studies validate the contribution of each proposed component, confirming that the combination of feature concatenation, multi-objective optimization, and VAD-Aware audio modeling provides a principled and effective solution for the EMI estimation task.
\section{Related Work}

The Affective Behavior Analysis in-the-Wild (ABAW) challenge has become one of the most influential benchmarks in affective computing, promoting robust understanding of human affect and behavior under unconstrained real-world conditions \cite{zafeiriou2017aff,kollias2019deep,kollias2020analysing,abaw2-competition,kollias2022abawcvpr,kollias2022abaweccv,kollias20247th,kollias2025advancements,kollias2025emotions}. Over the years, ABAW has continuously expanded its scope from core affective tasks such as valence-arousal estimation, expression recognition, and action unit detection to more complex settings involving multi-task learning, compound expressions, and broader behavior understanding \cite{kollias2019expression,kollias2019face,VA-expr-aus,kollias2021distribution,kollias2023multi,kollias2024distribution,kollias2024behaviour4all}.

In previous ABAW editions, the \textbf{Valence-Arousal (VA) Estimation} challenge focused on continuous prediction of affective dimensions in the wild \cite{zafeiriou2017aff,kollias2019deep,kollias2022abawcvpr}. The \textbf{Expression Recognition (EXPR)} challenge addressed categorical facial expression classification, while the \textbf{Action Unit (AU) Detection} challenge targeted fine-grained modeling of facial muscle movements \cite{kollias2019expression,kollias2019face,VA-expr-aus,kollias2022abawcvpr}. More recent ABAW challenges further extended the benchmark to richer affective and behavioral scenarios, including \textbf{compound expression recognition}, \textbf{violence detection}, and other forms of multimodal fine-grained behavior analysis \cite{kollias2023multi,kollias20247th,kollias2025dvd,kollias2025advancements,kollias2025emotions}.

In the 10th ABAW challenge, six tracks are organized, and the \textbf{Emotional Mimicry Intensity (EMI) Estimation} track is particularly distinctive. Unlike conventional affect recognition tasks that focus on predicting categorical labels or low-dimensional affective states, EMI aims to estimate how strongly a subject mimics emotionally evocative stimuli. This makes EMI a more subtle multimodal behavior understanding problem, requiring models to capture fine-grained intensity variations rather than coarse emotion categories.

The EMI track is built on the Hume-Vidmimic2 dataset and requires regression prediction over six continuous emotion dimensions, namely \textit{Admiration, Amusement, Determination, Empathic Pain, Excitement}, and \textit{Joy}. Compared with standard ABAW tasks, EMI is more challenging in several aspects. First, emotional mimicry is inherently multimodal, as it is jointly reflected through facial movements, vocal characteristics, and temporal coordination across modalities. Second, EMI is formulated as a continuous regression problem, which makes learning more sensitive to subtle label differences and optimization instability. Third, although the six target dimensions are semantically related, they also correspond to distinct behavioral patterns, raising the question of how to balance shared representation learning and dimension-specific discrimination.

Existing ABAW research has extensively studied unified affect modeling, multi-task optimization, and in-the-wild facial behavior analysis \cite{kollias2019expression,kollias2019face,VA-expr-aus,kollias2021distribution,kollias2024distribution}. These studies have shown the importance of jointly modeling related affective tasks and learning robust representations from large-scale unconstrained data. However, EMI estimation remains relatively underexplored and poses additional challenges due to its fine-grained regression nature and reliance on subtle multimodal mimicry cues. In this context, an important question is how to preserve modality-specific characteristics while still benefiting from multimodal fusion.
\begin{figure*}[htbp] 
    \centering 
    \includegraphics[width=1\textwidth]{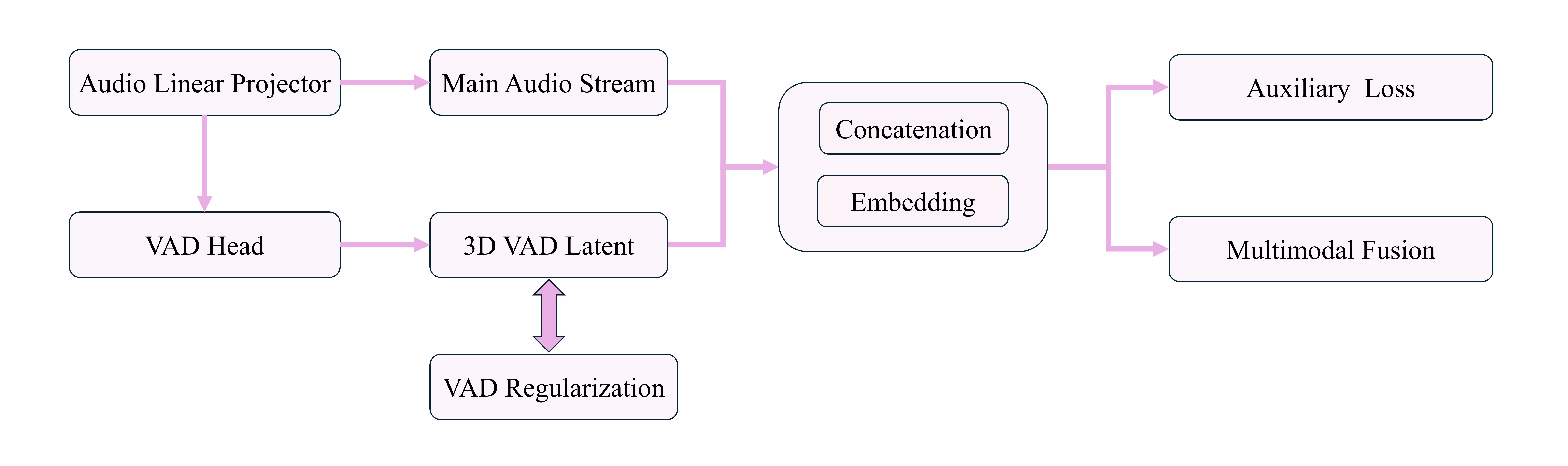} 
    \caption{Overview of the VAD-aware audio representation learning module.
After audio linear projection, the acoustic feature is split into a main audio stream and a VAD prediction branch. The VAD branch produces a 3D latent VAD representation, which is combined with the main audio stream to construct the audio embedding. This embedding is then used for multimodal fusion and auxiliary loss, while the latent VAD representation is regularized to improve stability and prevent extreme deviations.} 
    \label{fig:vad_audio_branch}
\end{figure*}
Motivated by this, we participate in the 10th ABAW EMI track and propose a simple yet effective multimodal regression framework. Through systematic exploration of pretrained high-level features, we find that direct feature concatenation can outperform more sophisticated fusion strategies for this task. Based on this empirical finding, our method is built upon three key principles: \textit{(i)} preserving modality-specific attributes through feature-level concatenation; \textit{(ii)} improving training stability and metric alignment via multi-objective optimization; and \textit{(iii)} enriching acoustic representations with a VAD-inspired latent prior. Our final framework integrates concatenation-based multimodal fusion, a shared six-dimensional regression head, multi-objective optimization, Exponential Moving Average (EMA) for parameter stabilization, and a VAD-inspired latent prior for the acoustic branch.

\section{Method}
\subsection{Problem Formulation}
Given a multimodal sample consisting of visual, audio, and text features, our goal is to predict a six-dimensional continuous target vector:
\begin{equation}
\mathbf{y} \in [0,1]^6,
\end{equation}
where each dimension corresponds to one EMI target.

\subsection{Overall Architecture}
As shown in Figure 1, the proposed framework adopts three modality-specific branches for visual, textual, and acoustic inputs. Each branch first applies a lightweight linear projector to pretrained features, producing compact modality representations $f_v$, $f_t$, and $f_a$. To improve optimization, auxiliary regressors are attached to the three branches to provide modality-level supervision. In addition, the acoustic branch is enhanced by a VAD-Aware latent pathway, which injects an affect-inspired prior into audio representation learning.

The final multimodal representation is formed by concatenating the projected visual, textual, and enhanced acoustic features, and is then passed to a shared six-dimensional regression head for EMI prediction. The model is trained with a multi-objective optimization scheme that combines the main regression loss, a Pearson-correlation objective, and auxiliary branch supervision. As verified by extensive ablation studies, this streamlined architecture achieves stronger and more stable performance than several complex variants. These observations suggest that, in our setting, stable feature-level fusion and optimization design are key to effective performance.

\subsection{Feature-Concat Multimodal Regression}
Let $\mathbf{z}_v$, $\mathbf{z}_a$, and $\mathbf{z}_t$ denote the visual, acoustic, and textual representations, respectively. Instead of directly averaging modality embeddings, we concatenate them:
\begin{equation}
\mathbf{z}_{fus} = [\mathbf{z}_v; \mathbf{z}_a; \mathbf{z}_t].
\end{equation}
The fused representation is passed to a shared regression head:
\begin{equation}
\hat{\mathbf{y}} = f_{reg}(\mathbf{z}_{fus}).
\end{equation}
This design preserves modality-specific heterogeneity and allows the regression head to learn cross-modal interactions adaptively.

\begin{table*}[htbp]
\centering
\caption{Pearson’s correlation for EMI Estimation on the Hume-Vidmimic2’s validation set \cite{table}.}
\label{table-Main} 
\resizebox{\textwidth}{!}{%
\begin{tabular}{cccccccccc}

\hline
Modality                       & Model                                         & Features               &  $\bar{p}$               & Admiration & Amusement & Determination & Empathic Pain & Excitement & Joy    \\ \hline
Faces                          & \multicolumn{2}{c}{Baseline ViT \cite{baseline}   }                              & 0.0900          & -          & -         & -             & -             & -          & -      \\
Audio                          & \multicolumn{2}{c}{wav2Vec2  \cite{baseline}}                                  & 0.2400          & -          & -         & -             & -             & -          & -      \\
Audio+Video                    & \multicolumn{2}{c}{ViT+wav2Vec2  \cite{baseline}}                              & 0.2500          & -          & -         & -             & -             & -          & -      \\
Audio+Video                    & \multicolumn{2}{c}{ResNet18+AUs+Wav2Vec2.0 \cite{bigtable43}}                 & 0.3288          & -          & -         & -             & -             & -          & -      \\
Audio                          & \multicolumn{2}{c}{Wav2Vec2.0+VAD \cite{bigtable8}}                               & 0.3890          & -          & -         & -             & -             & -          & -      \\
Audio+Video                    & \multicolumn{2}{c}{EmoViT+HuBERT+ChatGLM3 \cite{0.7}}                   & 0.5851          & 0.7155     & 0.6159    & 0.6303        & 0.3488        & 0.6174     & 0.5793 \\
\hline
\multirow{4}{*}{Faces}         & \multirow{4}{*}{MT-EmotiMobileViT}            & Embeddings (mean)      & 0.1644          & 0.0379     & 0.2314    & 0.1387        & 0.0781        & 0.2334     & 0.2672 \\
                               &                                               & Embeddings (STAT)      & 0.1683          & 0.0433     & 0.2459    & 0.1347        & 0.0779        & 0.2382     & 0.2699 \\
                               &                                               & Scores (mean)          & 0.1642          & 0.0321     & 0.2484    & 0.1490        & 0.0674        & 0.2399     & 0.2481 \\
                               &                                               & Scores (STAT)          & 0.1776          & 0.0619     & 0.2623    & 0.1247        & 0.0773        & 0.2544     & 0.2849 \\
\hline
\multirow{4}{*}{Faces}         & \multirow{4}{*}{MT-EmotiMobileFaceNet}        & Embeddings (mean)      & 0.1518          & 0.0215     & 0.2288    & 0.1140        & 0.0692        & 0.2299     & 0.2476 \\
                               &                                               & Embeddings (STAT)      & 0.1646          & 0.0557     & 0.2380    & 0.1303        & 0.0703        & 0.2325     & 0.2605 \\
                               &                                               & Scores (mean)          & 0.1667          & 0.0276     & 0.2367    & 0.1336        & 0.0807        & 0.2516     & 0.2699 \\
                               &                                               & Scores (STAT)          & 0.1732          & 0.0285     & 0.2498    & 0.1318        & 0.0970        & 0.2543     & 0.2776 \\
\hline
\multirow{2}{*}{Audio}         & \multirow{2}{*}{wav2vec 2.0}                  & Embeddings (mean)      & 0.1514          & 0.2153     & 0.1176    & 0.1834        & 0.1426        & 0.1275     & 0.1219 \\
                               &                                               & Embeddings (STAT)      & 0.2311          & 0.3006     & 0.1659    & 0.2559        & 0.3198        & 0.1844     & 0.1602 \\
\hline
\multirow{2}{*}{Audio}         & \multirow{2}{*}{HuBERT}                       & Embeddings (mean)      & 0.3045          & 0.3644     & 0.3179    & 0.2987        & 0.3111        & 0.2848     & 0.2499 \\
                               &                                               & Embeddings (STAT)      & 0.2729          & 0.3585     & 0.2454    & 0.2432        & 0.3065        & 0.2566     & 0.2275 \\
\hline
\multirow{2}{*}{Text}          & \multirow{2}{*}{RoBERTa (emotional)}          & Embeddings (mean)      & 0.3763          & 0.4558     & 0.3328    & 0.3427        & 0.4964        & 0.3201     & 0.3099 \\
                               &                                               & Embeddings (STAT)      & 0.3697          & 0.4501     & 0.3326    & 0.3173        & 0.4843        & 0.3331     & 0.3008 \\
\hline
\multirow{5}{*}{Text}          & SentenceBERT                                  & Embeddings (mean)      & 0.3756          & 0.4710     & 0.3585    & 0.3589        & 0.4347        & 0.3186     & 0.3119 \\
                               & Sentence all-MiniLM                           & Embeddings (mean)      & 0.3537          & 0.4552     & 0.3076    & 0.3371        & 0.4111        & 0.3093     & 0.3023 \\
                               & SentenceE5                                    & Embeddings (mean)      & 0.3777          & 0.4719     & 0.3497    & 0.3417        & 0.4553        & 0.3293     & 0.3184 \\
                               & OpenAI (small)                                & Embeddings (mean)      & 0.4011          & 0.5113     & 0.3825    & 0.3687        & 0.4847        & 0.3363     & 0.3233 \\
                               & GigaChat                                      & Embeddings (mean)      & 0.4001          & 0.4888     & 0.3648    & 0.3732        & 0.4862        & 0.3726     & 0.3151 \\
\hline
\multirow{2}{*}{Audio + Video} & \multicolumn{2}{c}{wav2vec 2.0 + MT-EmotiMobileViT}                    & 0.2829          & 0.3011     & 0.2968    & 0.2595        & 0.3074        & 0.3171     & 0.2152 \\
                               & \multicolumn{2}{c}{wav2vec 2.0 + MT-EmotiMobileFaceNet}                & 0.2898          & 0.3041     & 0.3004    & 0.2584        & 0.3148        & 0.3160     & 0.2452 \\
Text + Audio                   & \multicolumn{2}{c}{RoBERTa + HuBERT, blending}                         & 0.3974          & 0.4644     & 0.3727    & 0.3814        & 0.4755        & 0.3477     & 0.3430 \\
Text + Video                   & \multicolumn{2}{c}{RoBERTa + MT-EmotiMobileViT, blending}              & 0.4028          & 0.4379     & 0.3911    & 0.3580        & 0.4802        & 0.3656     & 0.3840 \\
Text + Video                   & \multicolumn{2}{c}{RoBERTa + MT-EmotiMobileFaceNet, blending}          & 0.4074          & 0.4408     & 0.3877    & 0.3632        & 0.5064        & 0.3650     & 0.3814 \\
Text + Video + Audio           & \multicolumn{2}{c}{RoBERTa + MT-EmotiMobileViT + HuBERT, blending}     & 0.4192          & 0.4603     & 0.4104    & 0.3844        & 0.4935        & 0.3802     & 0.3864 \\
Text + Video + Audio           & \multicolumn{2}{c}{RoBERTa + MT-EmotiMobileFaceNet + HuBERT, blending} & 0.4223          & 0.4620     & 0.4095    & 0.3848        & 0.4936        & 0.3876     & 0.3965 \\
\hline
Text + Audio                   & \multicolumn{2}{c}{OpenAI + HuBERT, blending}                          & 0.4125          & 0.5114     & 0.4006    & 0.3776        & 0.4863        & 0.3645     & 0.3344 \\
Text + Video                   & \multicolumn{2}{c}{OpenAI + MT-EmotiMobileViT, blending}               & 0.4103          & 0.4932     & 0.4090    & 0.3327        & 0.4571        & 0.3771     & 0.3930 \\
Text + Video                   & \multicolumn{2}{c}{OpenAI + MT-EmotiMobileFaceNet, blending}           & 0.3887          & 0.4779     & 0.3593    & 0.3369        & 0.4373        & 0.3452     & 0.3758 \\
Text + Video + Audio           & \multicolumn{2}{c}{OpenAI + MT-EmotiMobileViT + HuBERT, blending}      & 0.4451          & 0.5203     & 0.4461    & 0.3938        & 0.5054        & 0.4015     & 0.4036 \\
Text + Video + Audio           & \multicolumn{2}{c}{OpenAI + MT-EmotiMobileFaceNet + HuBERT, blending}  & 0.4225          & 0.5073     & 0.4044    & 0.3696        & 0.4801        & 0.3822     & 0.3912 \\
\hline
Text + Audio                   & \multicolumn{2}{c}{GigaChat + HuBERT, early}                           & 0.4011          & 0.4794     & 0.3733    & 0.4182        & 0.4449        & 0.3516     & 0.3395 \\
Text + Audio                   & \multicolumn{2}{c}{GigaChat + HuBERT, blending}                        & 0.4106          & 0.4955     & 0.3912    & 0.3881        & 0.4619        & 0.3922     & 0.3348 \\
Text + Video                   & \multicolumn{2}{c}{GigaChat+MT-EmotiMobileViT,blending}                           & 0.4103          & 0.4932     & 0.4090    & 0.3327       & 0.4571        & 0.3711     & 0.3930 \\
Text + Video                   & \multicolumn{2}{c}{GigaChat+MT-EmotiMobileFaceNet,early}                           & 0.4199          & 0.4906     & 0.3988    &  0.4121       & 0.4420        & 0.3876     & 0.3930 \\
Text + Video                   & \multicolumn{2}{c}{GigaChat+MT-EmotiMobileFaceNet,blending}                           & 0.4231          & 0.4897     &  0.4107    &   0.3806       &  0.4824        & 0.4131     &  0.3618 \\
Text + Video + Audio                   & \multicolumn{2}{c}{GigaChat+MT-EmotiMobileViT+HuBERT,blending}                           &  0.4460          &  0.5197     &  0.4491    &   0.3916       &   0.4981        & 0.4192     &  0.3983 \\
Text + Video + Audio                   & \multicolumn{2}{c}{GigaChat+MT-EmotiMobileFaceNet+HuBERT,early}                           &  0.4204          &   0.4823     &   0.4013    &    0.4267       &    0.4496        &  0.3765     &  0.3983 \\
Text + Video + Audio                   & \multicolumn{2}{c}{GigaChat+MT-EmotiMobileFaceNet+HuBERT,blending}                           &  0.4338          &   0.4955     &  0.4332    &    0.3923       &    0.4640        &  0.4282     &  0.3863 \\
\hline
Text+Audio+Face                & \multicolumn{2}{c}{ChatGLM3+WavLM-Large+ViT,blending}                   & \textbf{0.4673} & 0.5125
     & 0.4630
    & 0.4338
        & 0.5599
        & 0.4350
     &  0.3995\\
Text+Audio+Face                & \multicolumn{2}{c}{ChatGLM3+WavLM-Large+ViT,concat}                   & \textbf{0.4786} & 0.5242
     & 0.4709
    & 0.4415
        & 0.5846
        & 0.4531
     &  0.3971
 \\ \hline
\end{tabular}%
} 
\end{table*}

\subsection{VAD-Aware Audio Modeling}
For the acoustic branch, inspired by \cite{bigtable8}, we incorporate a VAD-aware latent pathway to augment audio features with an affect-inspired prior, as illustrated in Fig.~\ref{fig:vad_audio_branch}. Specifically, given the projected acoustic sequence, we first compute a global vector and then map it to a three-dimensional Valence-Arousal-Dominance (VAD) representation:
\begin{equation}
\hat{\mathbf{v}} = f_{vad}(\mathbf{a}_{\mathrm{mean}}), \qquad \hat{\mathbf{v}} \in [0,1]^3.
\end{equation}
The resulting three-dimensional latent representation is injected into the acoustic feature stream to modulate the final audio embedding. This design provides the acoustic branch with a low-dimensional affect-inspired prior, helping regularize audio representation learning and improving downstream multimodal regression.

\subsection{Multi-Objective Optimization}
The final training objective consists of four parts.

\noindent\textbf{MSE loss.}
The main regression objective is defined as
\begin{equation}
\mathcal{L}_{mse} = \frac{1}{6}\sum_{i=1}^{6}(\hat{y}_i-y_i)^2.
\end{equation}
It directly penalizes prediction error in the six-dimensional regression target.

\noindent\textbf{Pearson correlation loss.}
Since the challenge metric is based on Pearson correlation, we additionally optimize a batch-level correlation objective:
\begin{equation}
\mathcal{L}_{corr} = 1 - \mathrm{PCC}(\hat{\mathbf{y}}, \mathbf{y}).
\end{equation}
This term encourages consistency between prediction trends and ground-truth trends.

\noindent\textbf{Auxiliary branch supervision.}
Each modality branch is equipped with an auxiliary regressor for the same six-dimensional EMI target, and the corresponding losses are combined as
\begin{equation}
\mathcal{L}_{aux} = \lambda_v \mathcal{L}_v + \lambda_a \mathcal{L}_a + \lambda_t \mathcal{L}_t.
\end{equation}
This improves branch-level representation quality and prevents branch collapse.

\noindent\textbf{Latent prior regularization.}
For the acoustic branch, the VAD-inspired latent prior is regularized as
\begin{equation}
\mathcal{L}_{vad} = \|\hat{\mathbf{v}} - 0.5\|_2^2.
\end{equation}

The final loss is
\begin{equation}
\mathcal{L} = \mathcal{L}_{mse}
+ \lambda_{corr}\mathcal{L}_{corr}
+ \lambda_{aux}\mathcal{L}_{aux}
+ \lambda_{vad}\mathcal{L}_{vad}.
\end{equation}
During evaluation, we use exponential moving average (EMA) parameters for more stable validation performance.

\begin{table}[]
\centering
\caption{ Hume-Vidmimic2 partition statistics.}
\label{table-Dataset}
\begin{tabular}{ccc}
\toprule
Partition  & Duration & Samples \\
\hline
Train      & 15:07:03 & 8072    \\
Validation & 9:12:02  & 4588    \\
Test       & 9:04:05  & 4586    \\
\hline
All        & 33:23:10 & 17246   \\
\bottomrule
\end{tabular}
\end{table}

\section{Experiments}
\subsection{Dataset and Evaluation Metric}
We evaluate our method on the Emotional Mimicry Intensity (EMI) Estimation Challenge benchmark based on the Hume-Vidmimic2 dataset. Each sample in the dataset is annotated with six normalized emotion intensity labels within the range of $[0,1]$: Admiration, Amusement, Determination, Empathic Pain, Excitement, and Joy. As detailed in Table \ref{table-Dataset}, the benchmark comprises a total of 17,246 video clips with an aggregate duration of over 33 hours. The dataset is officially partitioned into three subsets: 8,072 samples for training, 4,588 for validation, and 4,586 for testing.

Following the challenge protocol, performance is measured by the average Pearson correlation coefficient across the six emotion categories:
\begin{equation}
\bar{p} = \sum_{i=1}^{6} \frac{p_i}{6}
\end{equation}
where $p_i$ ($i \in \{1,2,\cdots,6\}$) denotes the Pearson correlation coefficient for the $i$-th emotion category, calculated as
\begin{equation}
p_i = \frac{\mathrm{cov}(y_i,\hat{y}_i)}{\sqrt{\mathrm{var}(y_i)\mathrm{var}(\hat{y}_i)}}
\end{equation}
Here, $\mathrm{cov}(y_i,\hat{y}_i)$ denotes the covariance between the ground-truth and predicted values, while $\mathrm{var}(y_i)$ and $\mathrm{var}(\hat{y}_i)$ denote the variances of the ground-truth and predicted values, respectively.

\subsection{Implementation Details}
Our framework operates based on pre-extracted multimodal features. Visual features are derived from a ViT model pretrained on the AffectNethq and CASIA-web-face datasets, achieving a validation accuracy of 0.872340 in 20 epochs. Acoustic features are extracted using WavLM-Large. For the text modality, we utilize Whisper-Large to transcribe the audio; failed transcriptions in the training and validation sets (649 silent segments in total) are replaced with placeholders to maintain structural consistency, and text features are subsequently extracted via ChatGLM3-6B.

All modalities are temporally aligned to a fixed length of 128 via adaptive average pooling. The model uses a hidden/fused dimension of 256, dropout ratio of 0.2, and batch size of 32. Unless specified, we employ AdamW ($1\times10^{-4}$ initial learning rate, weight decay $1\times10^{-4}$), train for 30 epochs with cosine annealing, and apply early stopping (patience: 8). Other settings include mixed-precision training, gradient norm clipping (max norm: 1.0), and EMA (decay: 0.999) from epoch 1.

The final configuration uses feature concatenation, multi-objective optimization, and VAD-Aware audio modeling.

\begin{table}[t]
\centering
\caption{Comparison of average Pearson correlation coefficients on the validation set in the single-modality setting.}
\label{table-Single}
\begin{tabular}{lcccc}
\toprule
         & Visual & Audio & Text &   $\bar{p}$       \\
         \midrule
Baseline & Yes       &       &      & 0.09     \\
Ours      & Yes      &       &      & 0.092504 \\
Baseline &        & Yes     &      & 0.24     \\
Ours      &        & Yes     &      & 0.355879 \\
Ours      &        &       & Yes    & 0.464771 \\
\bottomrule
\end{tabular}
\end{table}

\begin{table}[t]
\centering
\caption{Ablation study of our method on the validation set.}
\label{table-Ablation}
\begin{tabular}{lccc}
\toprule
Method & Fusion & $\bar{p}$ \\
\midrule
Baseline & Average  & 0.467417 \\
Baseline & Concat  & 0.472604 \\
Baseline + Temporal Enhancement & Concat & \textbf{0.441066} \\
Baseline + Multi-Objective & Concat  & 0.476641 \\
Baseline + Multi-Objective+ VAD & Concat  & \textbf{0.478567} \\
\bottomrule
\end{tabular}
\end{table}

\subsection{Main Results}
Table \ref{table-Main} reports the validation performance of different unimodal and multimodal systems. Among unimodal settings, text-based features provide the strongest results, while visual-only performance remains limited. Audio features also contribute substantially and become more effective when combined with other modalities. These observations confirm that EMI estimation benefits from complementary multimodal information.

Our final model uses ChatGLM3 text features, WavLM-Large audio features, ViT visual features, concatenation-based fusion, multi-objective optimization, and VAD-Aware audio modeling, achieves an average Pearson correlation of 0.4786 on the validation split. Compared with average fusion, concatenation-based fusion yields better performance in our experiments, suggesting that preserving modality-specific information is beneficial under the current feature setting. Overall, the results show that carefully designed multimodal integration and optimization are crucial for strong EMI regression performance.

\subsection{Single-modal analysis}
Table \ref{table-Single} compares the average Pearson correlation coefficients of different unimodal settings on the validation set, leading to the following conclusions.

First, the visual-only setting remains weak. Compared to the baseline, our method only marginally improves the average Pearson correlation coefficient from 0.0900 to 0.0925, indicating that relying solely on visual information is insufficient for reliable EMI regression under the current feature settings.

Second, the audio-only setting exhibits significantly stronger capability. Our audio model achieves a score of 0.3559 compared to the baseline’s 0.2400, suggesting that under unimodal conditions, acoustic cues provide substantially more information for Emotional Mimicry Intensity estimation than visual cues.

Third, the text-only setting achieves the best unimodal performance, reaching an average Pearson correlation coefficient of 0.4648. This indicates that text representations contain highly discriminative emotional and semantic information for the EMI task, making it the strongest single modality in our framework. This finding validates the observations of \cite{0.7}.

Overall, Table \ref{table-Single} clearly demonstrates a hierarchy of modality capabilities under unimodal conditions: text is the strongest, followed by audio, while visual modality is relatively weak when used alone. These results provide a basis for subsequent multimodal design, suggesting that text should serve as the primary semantic support, audio provides crucial emotional complement, and visual modality offers additional but relatively limited auxiliary information.

\subsection{Ablation Study}
Table \ref{table-Ablation} presents the ablation results of our method on the validation set. We first compare two basic fusion strategies and observe that replacing average fusion with concat fusion improves the average Pearson correlation from 0.4674 to 0.4726. This result indicates that preserving modality-specific information is more effective than directly averaging heterogeneous multimodal representations.

We further investigate whether introducing additional temporal enhancement on top of the pretrained multimodal features is beneficial. In our setting, adding temporal enhancement leads to a substantial drop in performance, reducing the average Pearson correlation to 0.4411. This suggests that, under our feature-extraction pipeline, the pretrained representations are already sufficiently informative, and additional temporal modeling may introduce redundant transformations or optimization interference rather than useful temporal cues.

Building upon the concatenation baseline, we then introduce the proposed multi-objective optimization strategy, which combines the main regression loss, a Pearson-correlation objective, and auxiliary branch supervision. This improves the average Pearson correlation from 0.4726 to 0.4766, demonstrating that jointly optimizing the main regression objective together with metric-aligned and branch-level auxiliary objectives provides a more effective training signal.

Finally, incorporating the VAD-Aware audio modeling module further improves the result to 0.4786, achieving the best performance among all evaluated variants. This gain suggests that introducing a VAD-inspired latent prior into the acoustic branch is beneficial for multimodal EMI regression.

Overall, the ablation results lead to three conclusions: 1) concat fusion is superior to average fusion; 2) additional temporal enhancement is not helpful in the best-performing setting and may even be detrimental; and 3) the main performance gains come from multi-objective optimization and VAD-Aware audio modeling.

\section{Conclusion}
This paper presents a multimodal regression framework for the EMI track of the 10th ABAW Challenge. Our study systematically analyzes the roles of fusion strategies, multi-objective optimization, and VAD-Aware audio modeling under the condition of pretrained multimodal features. Experiments demonstrate that additional temporal modules can actually disrupt pretrained multimodal features, thereby degrading performance. Our method achieves an average Pearson correlation of 0.4786 on the official validation set.

{\small
\bibliographystyle{ieeenat_fullname}
\bibliography{main}
}
\end{document}